\documentclass[journal]{IEEEtran}
\usepackage{amsmath}
\usepackage{graphicx}
\usepackage{amsfonts}
\usepackage{array,cite}
\usepackage{amssymb}
\usepackage{helvet}
\usepackage{color}
\usepackage{rawfonts,amsfonts,psfrag,flushend}
\usepackage{bm}
\usepackage{subfig}
\usepackage{float,algorithm,algorithmic}
\usepackage{booktabs}
\usepackage{multirow}
\usepackage{caption}

\begin{document}
\title{Feature Learning Viewpoint of AdaBoost and a New Algorithm}
\author{Fei Wang, Zhongheng Li, Fang He, Rong Wang,~\IEEEmembership{Member,~IEEE}, Weizhong Yu and Feiping Nie$^*$
\IEEEcompsocitemizethanks{\IEEEcompsocthanksitem F. Wang, ZH. Li and W. Yu are with the National Engineering Laboratory for Visual Information Processing and Applications, and the School of Electronic and Information Engineering, Xi��an Jiaotong University, Xi��an, Shaanxi 710049, China (e-mail: wfx@mail.xjtu.edu.cn, lizhongheng2010@gmail.com and yuwz05@xjtu.edu.cn).
\IEEEcompsocthanksitem F. He is with the  Xi'an Research Institute of Hi-Tech, Xi'an, Shaanxi 710025, China (e-mail:fanghe1107@gmail.com).
\IEEEcompsocthanksitem Rong Wang and F Nie are with the Center for OPTical IMagery Analysis and Learning (OPTIMAL), Northwestern Polytechnical University, Xi'an, Shaanxi  710072, China (e-mail:wangrong07@tsinghua.org.cn and feipingnie@gmail.com).}
}


\vspace{-10mm}

\IEEEtitleabstractindextext{
\begin{abstract}
The AdaBoost algorithm has the superiority of resisting overfitting. Understanding the mysteries of this phenomena is a very fascinating fundamental theoretical problem. Many studies are devoted to explaining it from statistical view and margin theory. In this paper, we illustrate it from feature learning viewpoint, and propose the AdaBoost+SVM algorithm, which can explain the resistant to overfitting of AdaBoost directly and easily to understand. Firstly, we adopt the AdaBoost algorithm to learn the base classifiers. Then, instead of directly weighted combination the base classifiers, we regard them as features and input them to SVM classifier. With this, the new coefficient and bias can be obtained, which can be used to construct the final classifier. We explain the rationality of this and illustrate the theorem that when the dimension of these features increases, the performance of SVM would not be worse, which can explain the resistant to overfitting of AdaBoost. 
\end{abstract}

}
\maketitle
\IEEEdisplaynontitleabstractindextext

\section{Introduction}\label{Sec.1}
\IEEEPARstart{T}{he} Adaboost algorithm \cite{FREUND1997,Freund1996}, which learns a ``strong" classifier by voting the weighted predictions of a set of ``weak" learners (slightly better than random guessing) \cite{Reyzin2006}, is one of the most influential classification algorithms \cite{GAO20131,Caruana2006,wu2008}. The excellent performance has been demonstrated both on benchmark datasets and real application\cite{bauer1999,dietterich2000,viola2001rapid,GAO20131,wang2008}.

According to Occam's razor \cite{blumer1987}, when a classifier was trained too complex, the performance of it would be even worse rather than better. This phenomena is called overfitting, which means that the trained model is so adaptable to the training data that it would exaggerate the slight fluctuations in the training data, leading to poor generalization performance\cite{zhou2014large,GAO20131}.
However, the AdaBoost algorithm has the superiority of resisting overfitting, which has been observed by many researches\cite{breiman1998,drucker1996,quinlan1996bagging}. Understanding the mysteries of this phenomena about AdaBoost algorithm is a fascinating fundamental theoretical problem\cite{zhou2014large,GAO20131}. Many studies are devoted to explaining the success of AdaBoost, which can be divided into statistical view and margin theory\cite{zhou2014boosting}.

In the statistical view, great efforts were made to illustrate the success of AdaBoost algorithm. Friedman et al. \cite{friedman2000additive} utilized the well-known statistical principles, additive modeling and maximum likelihood, to understand this mysterious phenomenon. Besides, many boosting-style algorithms were proposed with optimizing the potential loss functions in a gradient decent way\cite{buhlmann2003boosting,mason2000boosting,GAO20131}. Inspired by this optimal method, some boosting-style algorithms and their variants that were consistent to Bayes's under different conditions were presented. \cite{bartlett2006convexity,bartlett2007adaboost,bickel2006some,breiman2000some,jiang2004process,lugosi2004bayes,GAO20131,mukherjee2013rate}. However, the biggest problem of the statistical view is that these algorithms do not explain well why AdaBoost is resistant to overfitting\cite{zhou2014large,GAO20131}.

The margin theory is another direction to solve this problem. Schapire et al.\cite{schapire1998boosting} were the first ones to use the margin theory to explain this overfitting phenomenon. Generally speaking, the margin of an example associated with the classifier is a measuring standard of the classification ability\cite{wang2008}. Schapire et al.\cite{schapire1998boosting} demonstrated that AdaBoost model can produce a good margin distribution which is the key to the success of AdaBoost. Soon after that, Breiman \cite{breiman1999prediction} put a doubt on this margin explanation. He proposed a boosting-type algorithm named arc-gv which directly maximizes the minimum margin for the generalization error of a voting classifier. In experiments, arc-gv can generate a larger minimum margin than AdaBoost, but it brought higher generalization error. Breiman concluded that neither the margin distribution nor the minimum margin has influence to the generalization error. Later, Reyzin and Schapire\cite{Reyzin2006} found that, amazingly, Breiman had not controlled model complexity well in the experiments. They repeated Brieman��s experiments using decision stumps with two leaves. The results showed that arc-gv was with larger minimum margin, but worse margin distribution. Therefore, a convincing explanation is urgently needed\cite{wang2008}.

To support the margin theory, Wang et al. \cite{wang2011refined} proved a bound in terms of
a new margin measure called the Equilibrium margin (Emargin). The Emargin bound was uniformly sharper than Breiman��s minimum margin bound. The results suggested that the minimum margin may be not crucial for the generalization error and a small empirical error at Emargin implied a smaller bound of the generalization error.  Gao and Zhou \cite{GAO20131}. proposed the $k$th margin bound to defend the margin theory against Breiman's doubt by a series mathematical derivation. This model was uniformly tighter than Breiman��s as well as Schapire��s bounds and considered the same factors as Schapire et al. and Breiman. Zhou et al. \cite{zhou2014large} proposed the Large margin Distribution Machine (LDM) to achieve a better generalization performance by optimizing the margin distribution. The margin distribution was characterized by the first-order statistic margin mean and second-order statistic variance. Then, the margin mean was tried to be maximum. At the same time, the margin variance was tried to be minimum. This method realized satisfactory results. However, completely explaining AdaBoost's resistance to overfitting is still difficult.

In this paper, we illustrate the resistant overfitting phenomena of AdaBoost from feature learning viewpoint and using the SVM classifier to explain it. SVM classifier is very useful for its clear principles and competitive accuracy\cite{Sentelle2016}\cite{Xu2015}\cite{yuan2012recent}\cite{chang2015libsvm}\cite{scholkopf2002learning}\cite{vapnik2013nature}\cite{joachims2006training}\cite{hou2014multiple}. We regard the results of base classifiers of AdaBoost as features and input them to SVM and explain the rationality of doing this. This means when the iterations of AdaBoost increase, the features' dimension increases. \textbf{We illustrate that the margin of SVM (not the margin of AdaBoost itself) will not be smaller when the features' dimension increase.} This implies that the performance of our AdaBoost+SVM model will improve when the iterations increase, which can directly and easily explain the resistant overfitting phenomena of AdaBoost rather than the complex proof. By the way, we also illustrate that the error rate of a binary classifier is always
not bigger than 0.5, which is not noticed by other researchers.

The rest of this paper is organized as follows. In Section~\ref{sec.2}, we have a briefly survey on the related work of the AdaBoost Algorithm. In Section~\ref{sec.3}, we present our AdaBoost+SVM model. In Section~\ref{sec.4}, we validate our methods on different datasets. In Section~\ref{sec.5}, we come to a conclusion.

\section{Related work}\label{sec.2}
In this section, we briefly review the general AdaBoost algorithm and the popular theoretical explanation to AdaBoost from the view of margin theory.
\subsection{AdaBoost Algorithm}
AdaBoost algorithm\cite{FREUND1997} is one of boosting classification algorithms which can boost a group of ��weak�� classifiers to a ��strong�� classifier. These algorithms usually first use a base classify algorithm whose classification ability is just better than random guessing to train a base classifier from the initial training samples. Then adjust the sample weight according to the result of the base classifier, which makes the samples that was classified incorrectly be paid more attention to. And then use the adjusted samples to train a next base learner. After iterations, weighted are added to these base learners to form  the final classifier. Next is the description of AdaBoost algorithm.

Let $\bm S=\{(x_1,y_1),\cdots,(x_i,y_i),\cdots,(x_n,y_n)\}$ denote the training samples set in the binary classification situation. $x_i \in  \bm X \subseteq \mathbb{R}^{n}$ is the $i$th instance. $y_i \in \bm Y=\{-1,+1\}$ is the class label associated with $x_i$. AdaBoost algorithm is based on the additive model, which is the linear combination of the base classifiers $h_t(x)$:
\begin{equation}\label{Eq.1}
f(x)=\sum_{t=1}^{T}\alpha_t h_t(x),
\end{equation}
where $t=\{ 1,\cdots ,T\}$ denotes the iteration number, $h_t(x)$ are the base classifiers trained from base classification algorithm $\mathfrak{L}$ whose classified ability is just better than random guessing and $\alpha_t$ are the weight coefficients.

In Eq.(\ref{Eq.1}), $h_t(x)$ are learned from base classification algorithm $\mathfrak{L}$ based on the training sample with weight vector $D_t$ at $t$ iteration.
\begin{equation}\label{Eq.2}
h_t(x): \bm X\rightarrow\{-1,+1\}.
\end{equation}

\begin{algorithm}[t]
\caption{AdaBoost algorithm}
\label{alg1}
\begin{algorithmic}
\STATE  \textbf{Input:}\\ \quad\quad Training set $S=\{(x_1,y_1),(x_2,y_2),\cdots,(x_n,y_n)\}$; \\ \quad\quad Base classification algorithm $\mathfrak{L}$ ; \\ \quad \quad Number of learning rounds $T$.\\
Initialize sample weight vector $D_1=(\underbrace{\frac{1}{n},\cdots ,\frac{1}{n},\cdots ,\frac{1}{n}}_{n})$ \\
\textbf{for} $t=1,\cdots, T$:\\
\begin{itemize}
\item[1.] Using the base classification algorithm $\mathfrak{L}$ and current weight $D_t$ to learn the base classifier $h_t(x)$ by minimizing the classification error $\epsilon_t$ defined in Eq. (\ref{Eq.4});
\item[2.] Calculating the  coefficient $\alpha_t$ based on Eq. (\ref{Eq.3});
\item[3.] Updating the weight $D_{t+1}$ by Eq. (\ref{Eq.6});\\
\end{itemize}
\textbf{end}\\
Combining the obtained $h_t(x)$ according to Eq. (\ref{Eq.10}) to complete the final classifier.
\STATE  \textbf{Output:}\\
\quad\quad The final classifier $F(x)$.\\
\end{algorithmic}
\end{algorithm}

The weight vector $D_t$ denotes the weight of each instance in $S$ at $t$ iteration. $D_1$ is composed as
\begin{equation}\label{Eq.5}
D_1(i)=\frac{1}{n}, i=1,2,\cdots,n,
\end{equation}
and $D_{t+1}$ are composed as

\begin{equation}\label{Eq.6}
D_{t+1}(i)=\frac{D_t(i)}{Z_t}\exp(-\alpha_t y_i h_t(x_i)), i=1,2,\cdots,n,
\end{equation}
where $Z_t$ are the normalization factors and are calculated as
\begin{equation}\label{Eq.9}
{Z_t}=\sum_{i=1}^{n}D_t(i) \exp(-\alpha_t y_i h_t(x_i)).
\end{equation}


From Eq.(\ref{Eq.6}) and Eq.(\ref{Eq.9}) we know that $D_{t+1}$ are adjusted from $D_{t}$. Thus the samples which are classified incorrectly in $h_t(x)$ will  have higher weights in $t+1$ iteration.

Given training set $S$ and sample weight $D_t$, the object of $h_t(x)$ is to minimizing the classification error $\epsilon_t$. $\epsilon_t$ is calculated as
\begin{equation}\label{Eq.4}
\epsilon_t=P[h_t(x_i)\neq y_i]=\sum_{i=1}^{n}D_t(i)I[h_t(x_i)\neq y_i],
\end{equation}
where $P[\cdot]$  denote the probability and $I[\cdot]$ denote the logic value.

In Eq.(\ref{Eq.1}), $\alpha_t$ measure the importance of $h_t(x)$ in the final classifier and are calculated in the following way.
\begin{equation}\label{Eq.3}
\alpha_t=\frac{1}{2}\ln(\frac{1-\epsilon_t}{\epsilon_t}).
\end{equation}

From Eq.(\ref{Eq.4}) and Eq.(\ref{Eq.3}) we can know that when $\epsilon_t<0.5$ , $\alpha_t>0$ . And $\alpha_t$ would increase with $\epsilon_t$ decrease. In fact, AdaBoost minimizes the exponential loss function in this process\cite{friedman2000additive}.

We can also notice that in AdaBoost algorithm, $\epsilon_t$ must smaller than 0.5 . In this condition, we make a remark which is not be noticed by others as follow:

 \textbf{Remark.} The error rate of a binary classifier is always not bigger than 0.5 .

 \emph{Explanation.} For binary classification problems, if the error rate $\epsilon$ of weak classifier

\begin{equation}
h\left( x \right)=\left\{ \begin{matrix}
   1 & x\in \Omega   \\
   -1 & x\notin \Omega   \\
\end{matrix} \right.
\end{equation}
 is bigger than 0.5, we can use the classifier
\begin{equation}
h_c\left( x \right)=\left\{ \begin{matrix}
   -1 & x\in \Omega   \\
   1 & x\notin \Omega   \\
\end{matrix} \right.
\end{equation}
to replace $h(x)$, which makes the error rate $\epsilon_c$ convert to $1-\epsilon$. Then the $\epsilon_c$ is smaller than 0.5 .

\textbf{This remark illustrates that $\epsilon_t$ would always be smaller than 0.5 } unless equal to 0.5 . But it is almost impossible that the error rate of a classifier happens to be 0.5 .

After continuous iteration, the final classifier is
\begin{equation}\label{Eq.10}
F(x)= sign(f(x))=sign(\sum_{t=1}^{T}\alpha_t h_t(x)).
\end{equation}

This algorithm is summarized in Algorithm \ref{alg1}.


\subsection{Intuition of the Margin Theory}
AdaBoost is one of the most influential and successful classification algorithms. However,  the mystery of the phenomenon of its resistant overfitting attracted many scholars working on it. A theory which is intuitive to explain this phenomenon is the margin theroy. That is although the training error reaches zero, the margin of AdaBoost will increases along with the iterations increase.

Schapire et al.\cite{schapire1998boosting} were the first ones to use the margin theory to explain this overfitting phenomenon. Define $yf(x)$ as the margin for $(x,y)$ with respect to $f$. Use $P_D[\cdot]$ to refer as the probability with respect to sample weight vector $D$, and $P_S[\cdot]$ to denote the probability with respect to uniform distribution over the sample $S$. They first proved the following theorem to bound the generalization error of each voting classifier:

\textbf{Theorem 1}\ Let $S$ be a sample of $n$ examples chosen independently at random according to $D$. Assume that the base hypothesis space $\mathcal{H}$ and $\delta>0$. Then with probability at least $1-\delta$ over the random choice of the training set $S$, every weighted averange function $f\in\mathcal{C}$ satisfies the following bound for all $\theta >0$:

\begin{equation}\label{Eq.T1}
\begin{split}
{{P}_{D}}&\left[ yf\left( x \right)\le 0 \right]\le {{P}_{S}}\left[ yf\left( x \right)\le \theta  \right]+ \\
&O\left( \frac{1}{\sqrt{n}}{{\left( \frac{\log n\log \left| \mathcal{H} \right|}{{{\theta }^{2}}}+\log \frac{1}{\delta } \right)}^{1/2}} \right).
\end{split}
\end{equation}
If $\mathcal{H}$ is finite, then
\begin{equation}\label{Eq.T1_2}
\begin{split}
{{P}_{D}}&\left[ yf\left( x \right)\le 0 \right]\le {{P}_{S}}\left[ yf\left( x \right)\le \theta  \right]+ \\
&O\left( \frac{1}{\sqrt{n}}{{\left( \frac{d\log ^2(n/d)  }{{{\theta }^{2}}}+\log \frac{1}{\delta } \right)}^{1/2}} \right),
\end{split}
\end{equation}
where $d$ is the VC dimension of $\mathcal{H}$.

This theorem illustrates that if a voting classifier generates a good margin distribution, then the generalization error is also small.

Then they propose that if $\theta$ is not too large, the fraction of training examples for which $yf(x)\leq \theta$ decreases to zero exponentially fast with the number of base hypotheses.

\textbf{Theorem 2}\ Suppose the base learning algorithm, when called by AdaBoost, generates hypotheses with weighted training errors $\epsilon_1,\cdots,\epsilon_T$. Then for any $\theta$,
\begin{equation}\label{Eq.T2_1}
{{P}_{S}}\left[ yf\left( x \right)\le \theta  \right]\leq 2^T\prod_{t=1}^{T}\sqrt{\epsilon_t^{1-\theta}(1-\epsilon_t)^{1+\theta}}.
\end{equation}
Assume that, for all t, $\epsilon_t\leq 1/2-\gamma$ for some $\gamma>0$, the upper bound in Eq. \ref{Eq.T2_1} can simplify to:

\begin{equation}\label{Eq.T2_2}
\left(\sqrt{(1-2\gamma)^{1-\theta}(1+2\gamma)^{1+\theta}}\right)^T.
\end{equation}
If $\theta <\gamma$, it is easy to find that the expression inside the parentheses is smaller than 1 so that the probability that $yf(x)<\theta$ decreases exponentially fast with $T$. That is to say with the $T$ increase, AdaBoost can provide better margin distribution, which seems to explained the resistant overfitting phenomenon. This explanation is quite intuitive.

Later, Breiman\cite{breiman1999prediction} proved a generalization bound, which is tighter than Eq.(\ref{Eq.T1}) and designed the arc-gv algorithm which directly maximizes the minimum margin. According to margin theory, arc-gv should perform better than AdaBoost. However, the experiments results show that arc-gv does produce uniformly larger minimum margin but the test error increases. Thus Breiman concluded that the margin theory was in serious doubt.

Several years later, Reyzin and Schapire\cite{Reyzin2006} found that Breiman had not controlled model complexity well in the experiments. They repeated Brieman��s experiments using decision stumps with two leaves. The results showed that arc-gv was with larger minimum margin, but worse margin distribution. Later, Wang et al. \cite{wang2011refined} proved a bound in terms of a new margin measure called the Equilibrium margin (Emargin). The Emargin bound was uniformly sharper than Breiman��s minimum margin bound. The results suggested that the minimum margin may be not crucial for the generalization error and a small empirical error at Emargin implied a smaller bound of the generalization error.  Gao and Zhou \cite{GAO20131}. proposed the $k$th margin bound to defend the margin theory against Breiman's doubt by a series mathematical derivation. This model was uniformly tighter than Breiman��s as well as Schapire��s bounds and considered the same factors as Schapire et al. and Breiman. Zhou et al. \cite{zhou2014large} proposed the Large margin Distribution Machine (LDM) to achieve a better generalization performance by optimizing the margin distribution. The margin distribution was characterized by the first-order statistic margin mean and second-order statistic variance. Then, the margin mean was tried to be maximum. At the same time, the margin variance was tried to be minimum. This method realized satisfactory results. However, completely explaining AdaBoost's resistance to overfitting is still difficult.

In fact, the certain relationship of the margin $yf(x)$ of AdaBoost itself with the iteration number $T$ is still not clearly from the works above. That is these works can not directly explain the resistant overfitting of AdaBoost when the iteration number $T$ increases even after the training error reaches $0$. In next section, we will introduce our AdaBoost+SVM model, which can give the certain relationship of iteration number $T$ and the SVM margin to explain the resistant overfitting phenomena of AdaBoost directly.

\begin{figure*}[!hbtp]
\centering
\includegraphics[scale=0.5]{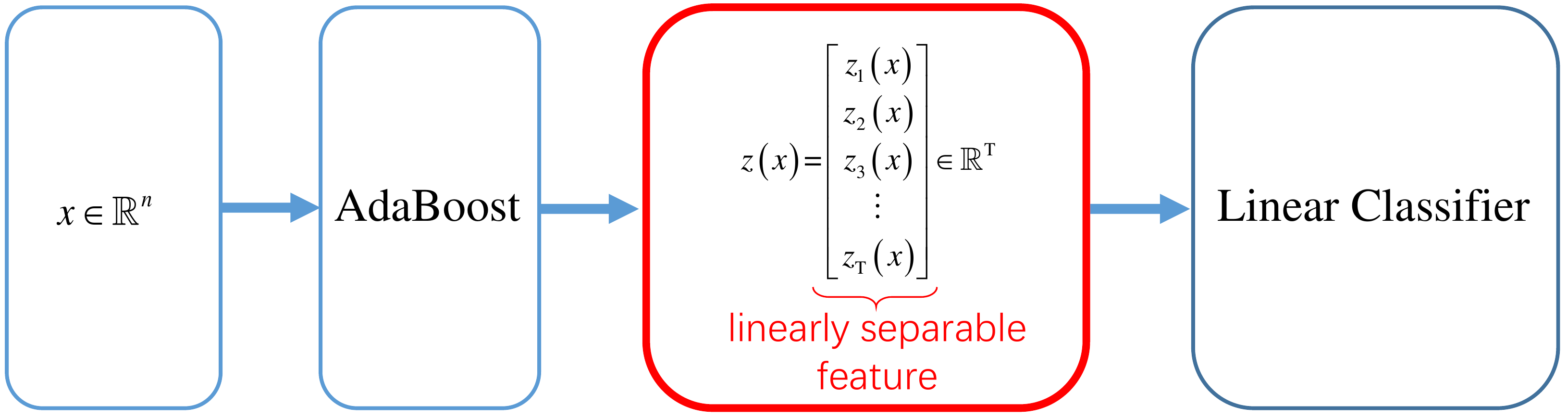}
\caption{The feature learning view of AdaBoost.\label{Fig.2}}
\end{figure*}

\section{Feature Learning Viewpoint}\label{sec.3}
In this section, we propose our AdaBoost+SVM model to explain the resistant overfitting phenomena of AdaBoost from the feature learning viewpoint and explain
the rationality of doing this. 

Freund and Schapire\cite{FREUND1997} have proved that the training error of AdaBoost decreases exponentially fast constantly during the learning process. There is a theorem as follows:

%

\textbf{Theorem 3} The training error of AdaBoost will always reach $0$ since the iterations increase.

This theorem comes from the following equation:

\begin{equation}
\frac{1}{n}\sum\limits_{i=1}^{n}{I\left( F\left( {{x}_{i}} \right)\ne {{y}_{i}} \right)}\le \exp \left( -2\sum\limits_{t=1}^{T}{\gamma _{t}^{2}} \right),
\end{equation}
where ${{\gamma }_{t}}=0.5-{{\epsilon}_{t}}$, $T$ denotes the number of iteration.

We have explained that $\epsilon_t$ is always smaller than 0.5 in previous chapter. Therefore $0<\gamma_t\leq0.5$ .

Let $\forall {\gamma }_{t}$, $0<\gamma\leq{{\gamma }_{t}}$,
\begin{equation}\label{Eq.22}
\frac{1}{n}\sum\limits_{i=1}^{n}{I\left( F\left( {{x}_{i}} \right)\ne {{y}_{i}} \right)}\le \exp \left( -2T{\gamma^{2}} \right).
\end{equation}

Then from Eq.(\ref{Eq.22}) we can easily know that the error of AdaBoost will be reduced at an exponential rate and always reach 0 when iterations are enough. Based on this theorem, we will propose our model to view AdaBoost from the feature learning point next.

\subsection{AdaBoost+hard margin SVM}
The boosting part $\sum\limits_{t=1}^{T}{{{\alpha }_{t}}{{h}_{t}}\left( x \right)}$ in final classifier of AdaBoost given in equal (\ref{Eq.10}) can be rewritten as
\begin{equation}\label{Eq.23}
f\left( x \right)=\left[ {{\alpha }_{1}},{{\alpha }_{2}},\cdots ,{{\alpha }_{T}} \right]\left[ \begin{matrix}
   {{h}_{1}}\left( x \right)  \\
   {{h}_{2}}\left( x \right)  \\
   \vdots   \\
   {{h}_{T}}\left( x \right)  \\
\end{matrix} \right].
\end{equation}
In Eq.(\ref{Eq.23}), we can regard the right vector
\begin{equation}\label{Eq.24}
z(x) = {{\left[ {{h}_{1}}\left( x \right),{{h}_{2}}\left( x \right),\cdots ,{{h}_{T}}\left( x \right) \right]}^{T}}
\end{equation}
as a feature of sample $x$ that learning from AdaBoost. Then $\alpha = \left[ {{\alpha }_{1}},{{\alpha }_{2}},\cdots ,{{\alpha }_{T}} \right]$ is the weight vector of this feature provided by AdaBoost algorithm. In other words, we can regard the process $x\rightarrow z(x)$ as a $\mathbb{R}^n\rightarrow \mathbb{R}^T $ spatial mapping. From this view, $\alpha$ can be viewed as a hyperplane in the feature space $\mathbb{R}^T$ and divide the features into two categories.

However, $\alpha$ may not be the best hyperplane for classifying the features. According to Theorem 2, the training error of AdaBoost will always reach $0$. And the training error of AdaBoost reaches $0$ means the feature $z(x)\in \mathbb{R}^T$ can be linearly separated into two categories. Another fact is that SVM algorithm can provide the separating hyperplane with largest margin in linearly separable problem, which means an excellent solution\cite{Wu2018}. Based on this, use SVM algorithm to calculate the hyperplane in the feature space $\mathbb{R}^T$ to replace $\alpha$ should be a better choice. This is the  theoretical basis of our model. The algorithm will be described next.

Given the training sample $S$, get the feature function $z(x)$ by AdaBoost algorithm according to Eq.(\ref{Eq.24}), first. Then, according to the following objective function of SVM:
\begin{align}\label{Eq.25}
& \min_{\beta, b}\, \frac{1}{2}\|\beta\|^2 \nonumber\\
& s.t.\quad y_i( \beta z(x_i)+b)-1\geq 0, i=1,2,\cdots,n,
\end{align}
we can calculate the optimal weight vector $\beta$ and bias $b$. Last, the final classifier can be learned by
\begin{equation}\label{Eq.26}
F(x)=sign(f(x))=sign(\sum_{t=1}^{T}\beta_t h_t(x)+b).
\end{equation}

We illustrate the feature learning view of AdaBoost in Fig.~\ref{Fig.2}.

\subsection{AdaBoost+soft magin SVM}
Although the training error of AdaBoost will always reach 0 with the iterations growing, in practical situation, the training error may usually not reach 0 beacause of the fixed iterations or large-scale data \emph{et al}. In this situation, the features from $z(x)\in \mathbb{R}^T$ maybe can not be linearly separated so that the hard margin SVM is not suitable. To solve this problem, we use soft margin SVM to replace the hard margin SVM, i.e. add an additional margin violation $\xi_i$ to Eq.(\ref{Eq.25}). The objective function is\cite{Chen2004}:
\begin{align}\label{Eq.27}
\underset{\beta ,b,\xi }{\mathop{\min }}\,& \frac{1}{2}{{\left\| \beta  \right\|}^{2}}+C\sum\limits_{i=1}^{n}{{{\xi }_{i}}}  \nonumber \\
s.t.\ &y_i( \beta z(x_i)+b)\geq 1-{\xi }_{i} \nonumber \\  
& {\xi }_{i} \geq 0, i=1,2,\cdots,n,
\end{align}
where $C$ represents the tolerance of the margin violation $\xi_i$.

Then we can use this way to solve our model whether the training error of AdaBoost reachs 0 or not.

The algorithm of AdaBoost+SVM is described in Algorithm \ref{alg3}.
\begin{algorithm}[t]
\caption{AdaBoost+SVM algorithm.}
\label{alg3}
\begin{algorithmic}
\STATE  \textbf{Input:}\\ \quad\quad Training sample $S$; \\ \quad\quad Base learning algorithm $\mathfrak{L}$; \\ \quad \quad Number of base learners $T$.\\
\begin{itemize}
\item[1.] Using AdaBoost to get the feature function $z(x)$ \\
according to Eq.(\ref{Eq.24});
\item[2.] Using SVM classifier to calculate the new coefficient $\beta$ and the bias $b$ according to Eq.(\ref{Eq.27});
\item[3.] Combining the obtained $\beta$ , $b$  and the feature function  to complete the final classifier according to Eq. (\ref{Eq.26}).
\end{itemize}

\STATE  \textbf{Output:}\\
\quad\quad The final classifier $F(x)$.\\
\end{algorithmic}
\end{algorithm}

\subsection{The explanation to the resistant overfitting by our model}
Overfitting is a common problem in many classification situation. However, the AdaBoost algorithm can resist overfitting. Understanding the mystery is a very fascinating fundamental theoretical problem. Our AdaBoost+SVM model also has the superiority of explaining this phenomenon. We utilize the following theorem to analyze this property from feature learning viewpoint.

\textbf{Theorem 4} \ SVM is a linear classifier in the feature space. As the dimensions of features increases, the margin of SVM will not be smaller.

\emph{Proof.} The objective function of SVM is:
\begin{align}\label{Eq.16}
&\min_{\beta, b}\,\frac{1}{2}\|\beta\|^2 \nonumber\\
&s.t.\quad y_i( \beta x_i+b)-1 \geq 0, i=1,2,\cdots,n.
\end{align}

The optimal solution $\beta^*$ and $b^*$ can be calculated. Then, the corresponding
hard margin separation hyperplane is
\begin{equation}\label{Eq.17}
\beta^*x+b^*=0.
\end{equation}

If $x$ increases to $(x, x_t)$, then the corresponding $\beta$ becomes to $(\beta, \beta_t)$.
The new optimal solution $\beta_{new}^*$ and $b_{new}^*$ can be obtained. Then, the corresponding hard margin separation hyperplane is
\begin{equation}\label{Eq.18}
\beta_{new}^*x+b_{new}^*=0.
\end{equation}

If $\beta_t=0$, the margin will stay the same. If $\beta_t\neq 0$, the margin will become larger. In a word, the margin does not become smaller when the dimensions of features are increased. According to\cite{zhou2014boosting}, the bigger margin is, the higher the predictive confidence is. Therefore, as the number of feature increases, the classification performance will not decrease.

Based on this theorem, we regard the obtained results of base classifiers by the AdaBoost algorithm as features of SVM. As the dimensions of features increases, the performance of SVM classifier would be improved, which can easily explain the advantage of AdaBoost that can resist overfitting. Therefore, our AdaBoost+SVM model have illustrated the mysteries of resistant overfitting from the feature learning viewpoint.

It should be noticed that as the number of $T$ increased, we can not directly obtain the $\theta$ in Eq.(\ref{Eq.T2_1}) also increase. But in our model, the increase of $T$ must be helpful for better performance, which can explain the resistant overfitting phenomena of AdaBoost directly and easily.

\section{Experiment}\label{sec.4}
In this section, we conduct experiments on four binary benchmark datasets to demonstrate the efficiency and effectiveness of the proposed method. Then, we have a detail analysis about the experimental results.

\begin{figure}[!hbtp]
\centering
\subfloat[]{\includegraphics[scale=0.5]{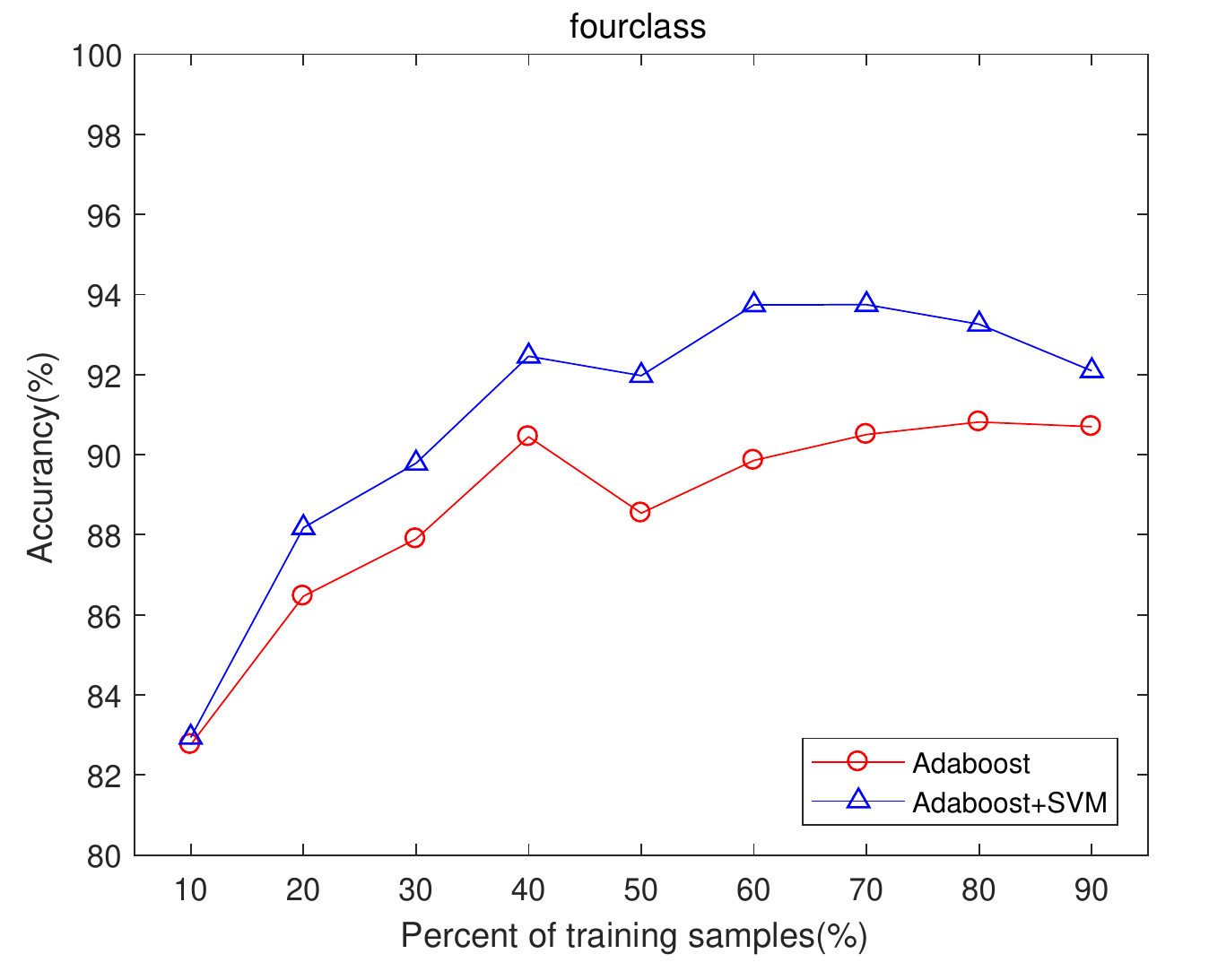}\label{Fig.3a}}\
\subfloat[]{\includegraphics[scale=0.5]{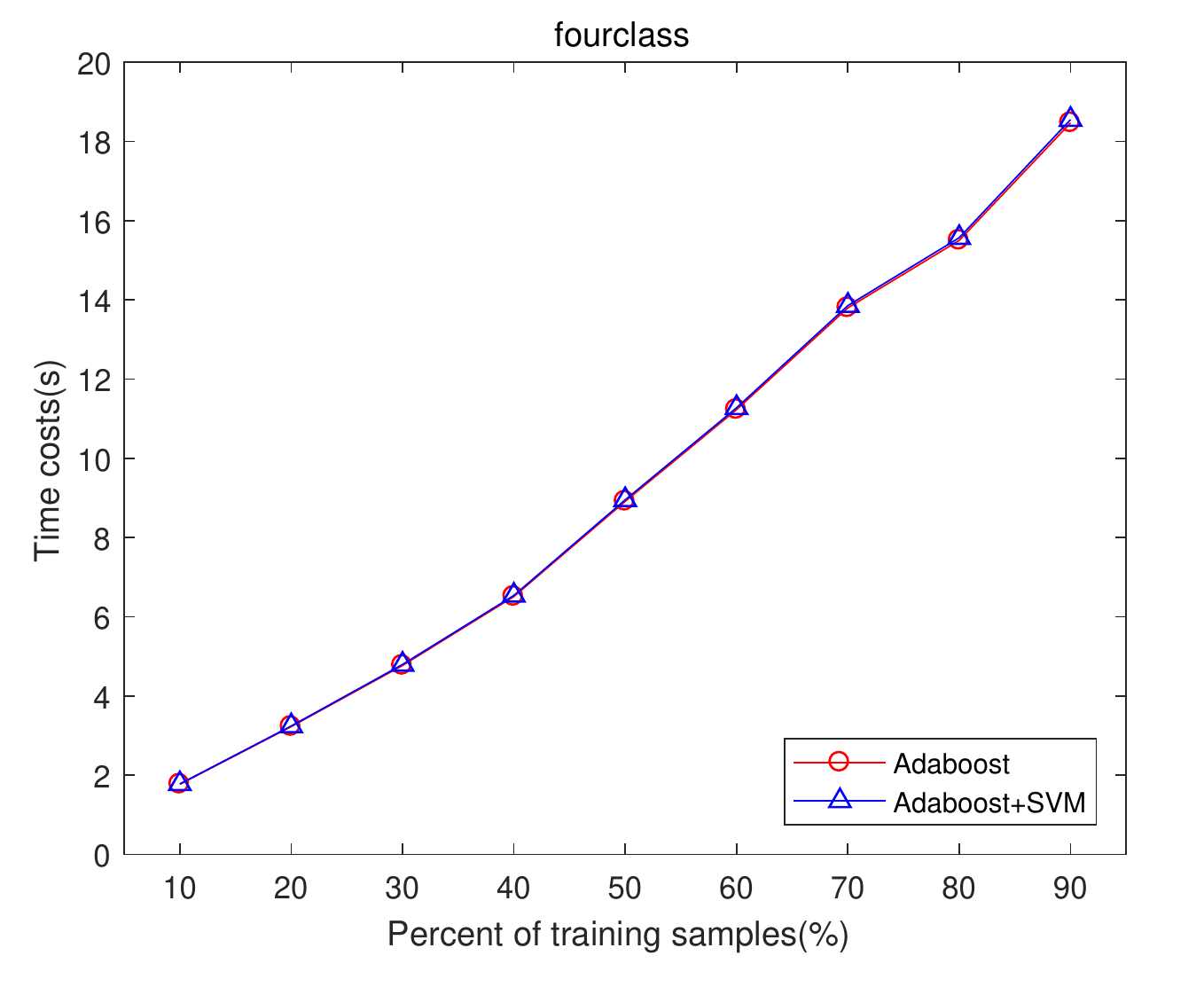}\label{Fig.3b}}\\
\caption{Classification accuracy ($\%$) and time costs vs the percent of labeled samples on fourclass dataset.\label{Fig.3}}
\end{figure}
\begin{figure}[!hbtp]
\centering
\subfloat[]{\includegraphics[scale=0.5]{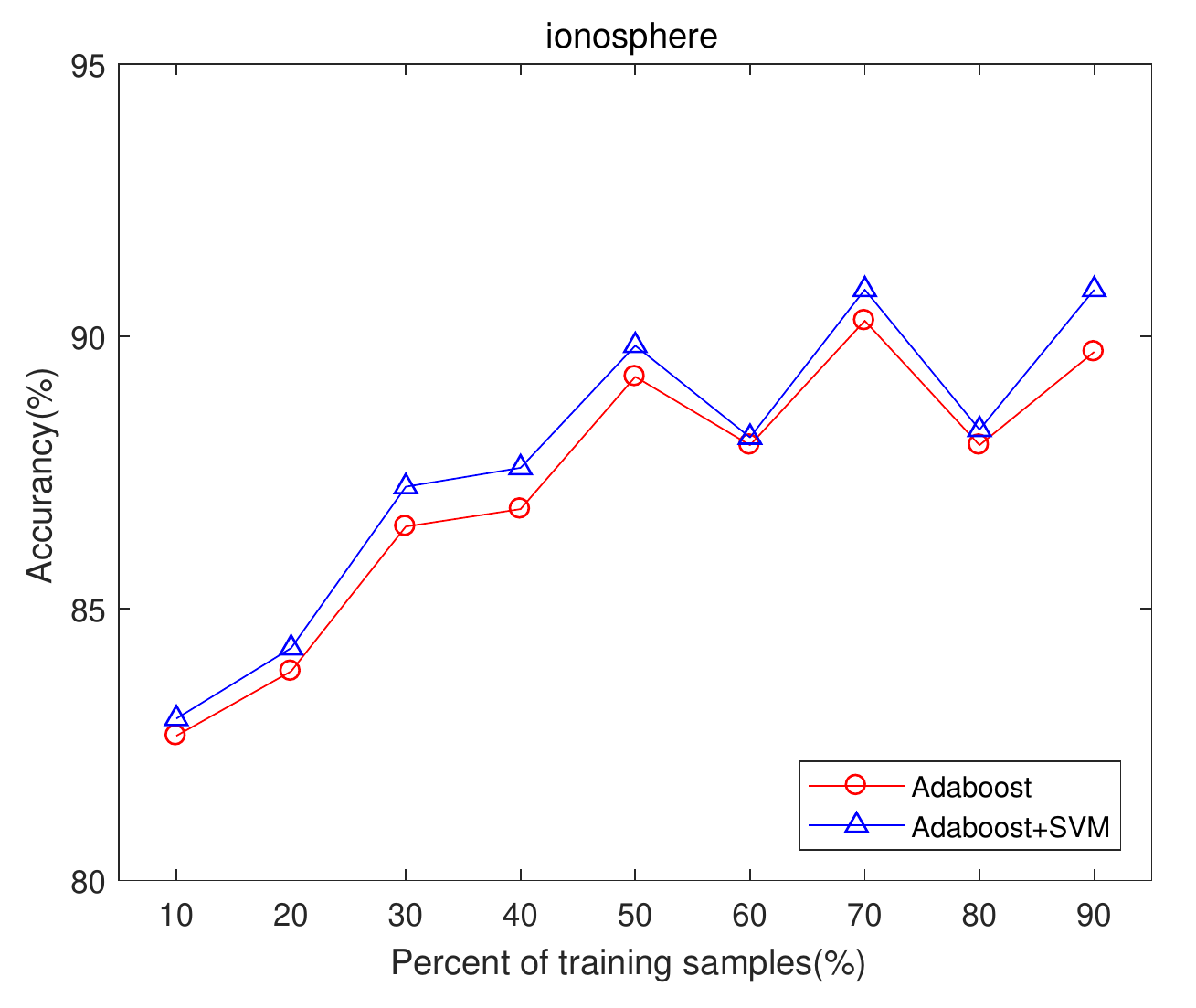}\label{Fig.5a}}\
\subfloat[]{\includegraphics[scale=0.5]{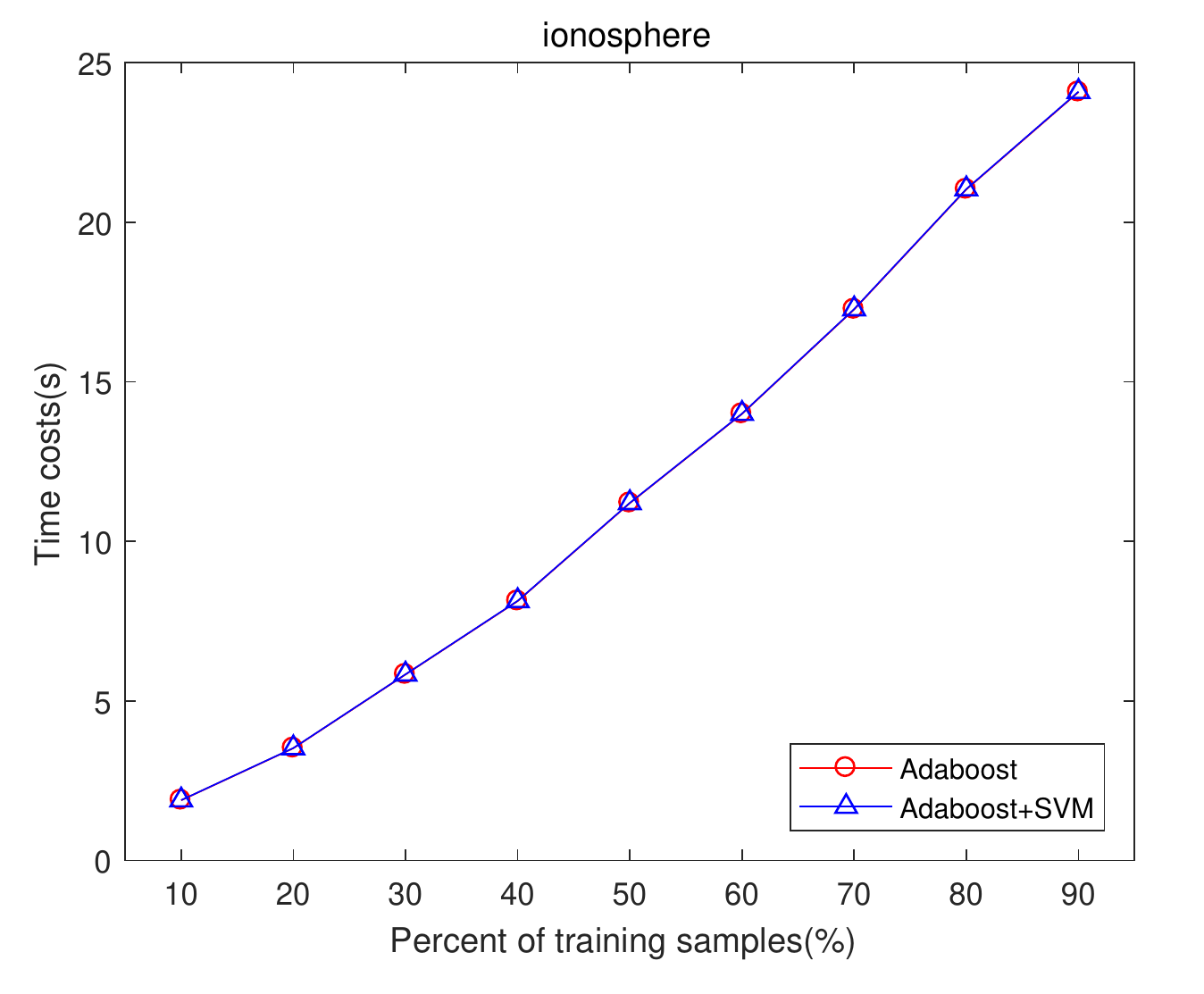}\label{Fig.5b}}\\
\caption{Classification accuracy ($\%$) and time costs vs the percent of labeled samples on ionosphere dataset.\label{Fig.4}}
\end{figure}

\begin{figure}[!hbtp]
\centering
\subfloat[]{\includegraphics[scale=0.5]{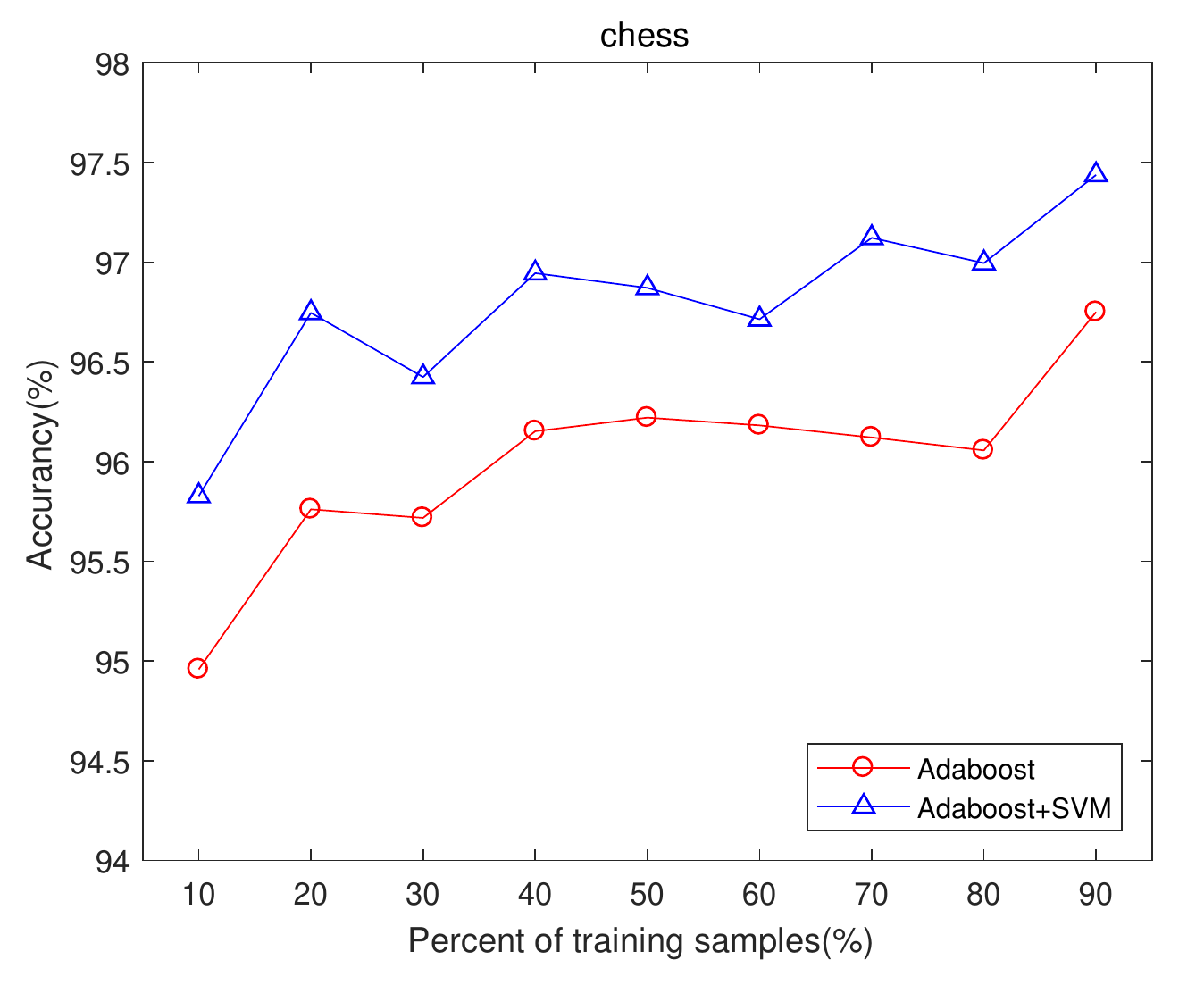}\label{Fig.6a}}\
\subfloat[]{\includegraphics[scale=0.5]{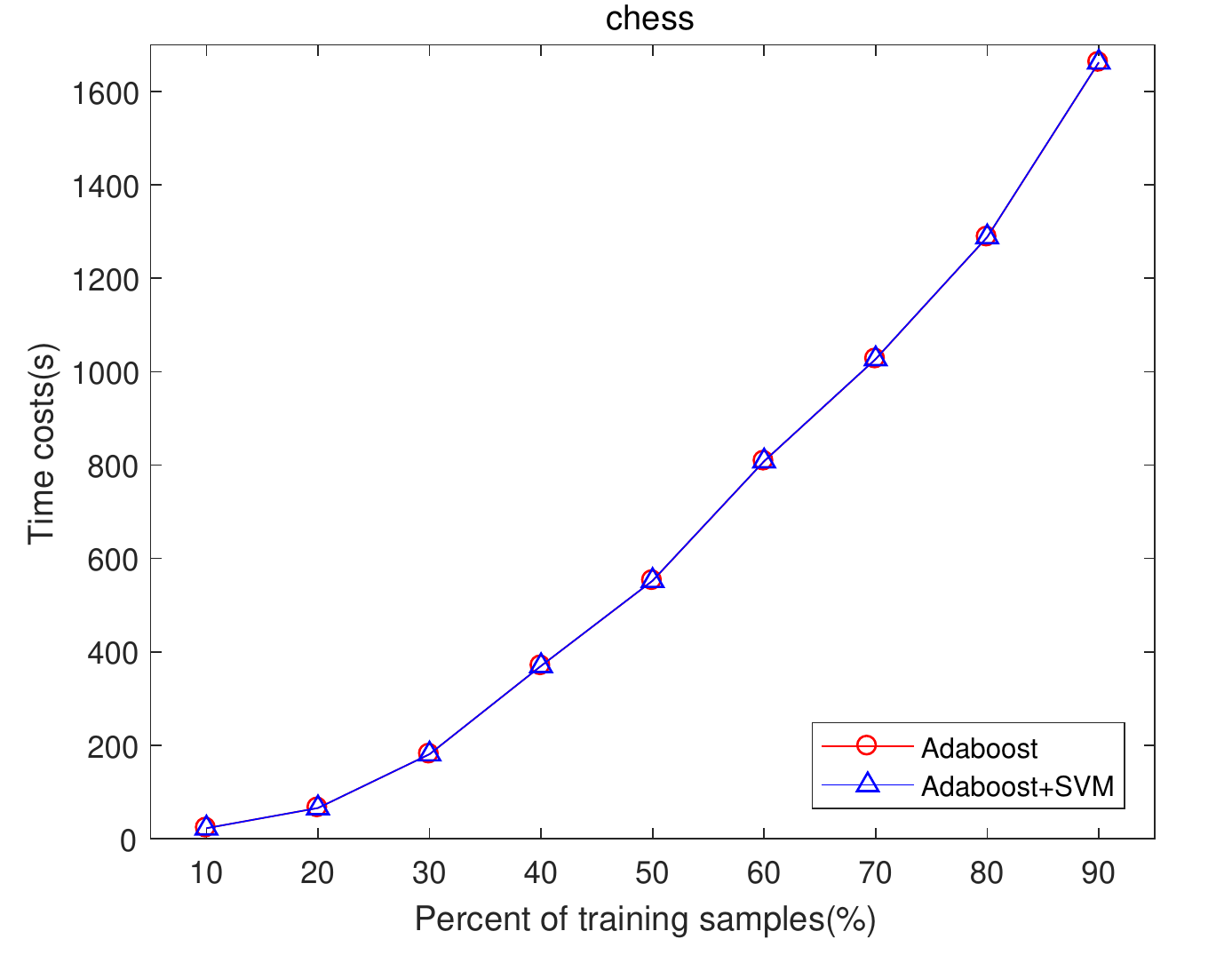}\label{Fig.6b}}\\
\caption{Classification accuracy ($\%$) and time costs vs the percent of labeled samples on chess dataset.\label{Fig.5}}
\end{figure}

\begin{figure}[!hbtp]
\centering
\subfloat[]{\includegraphics[scale=0.5]{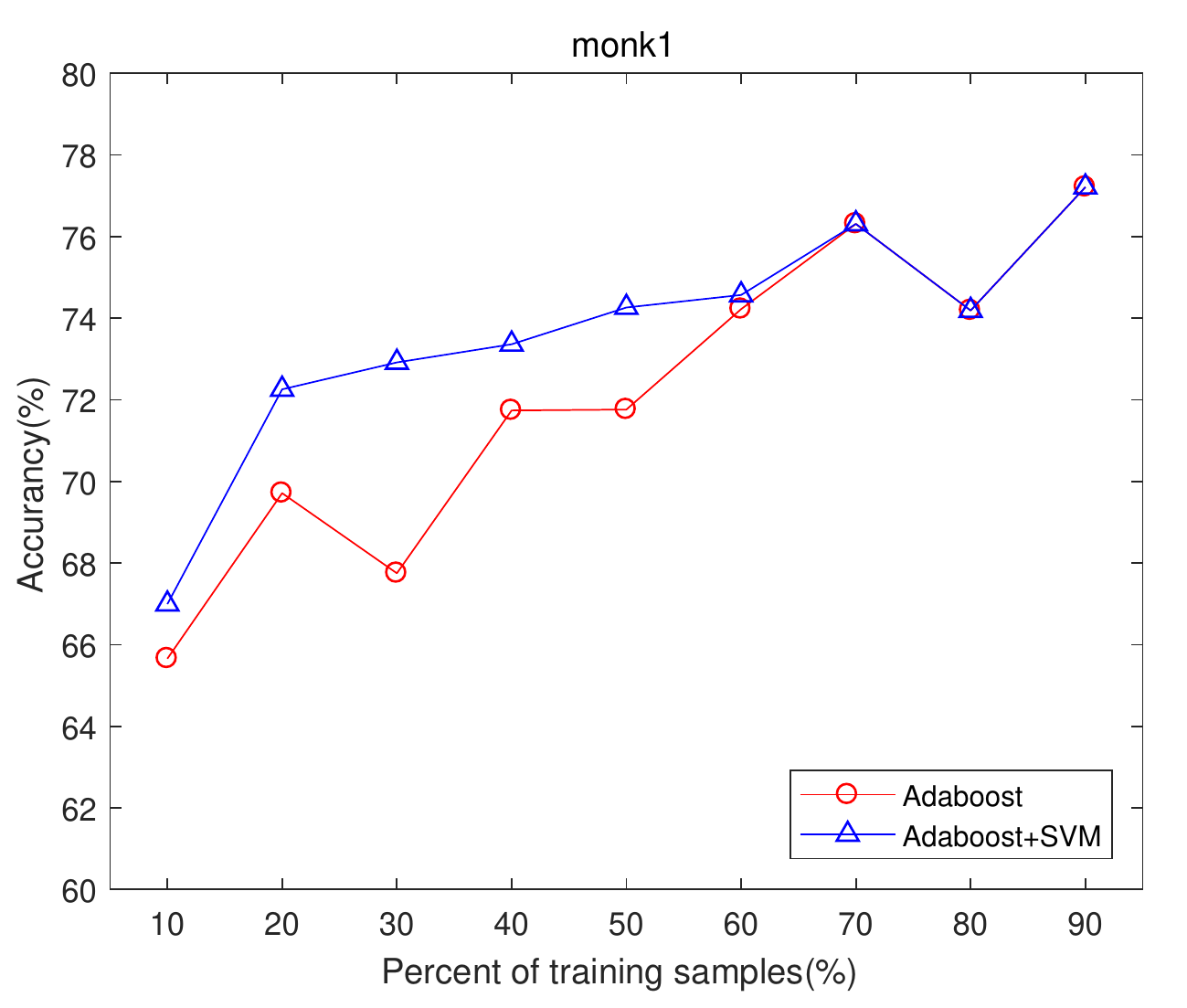}\label{Fig.8a}}\
\subfloat[]{\includegraphics[scale=0.5]{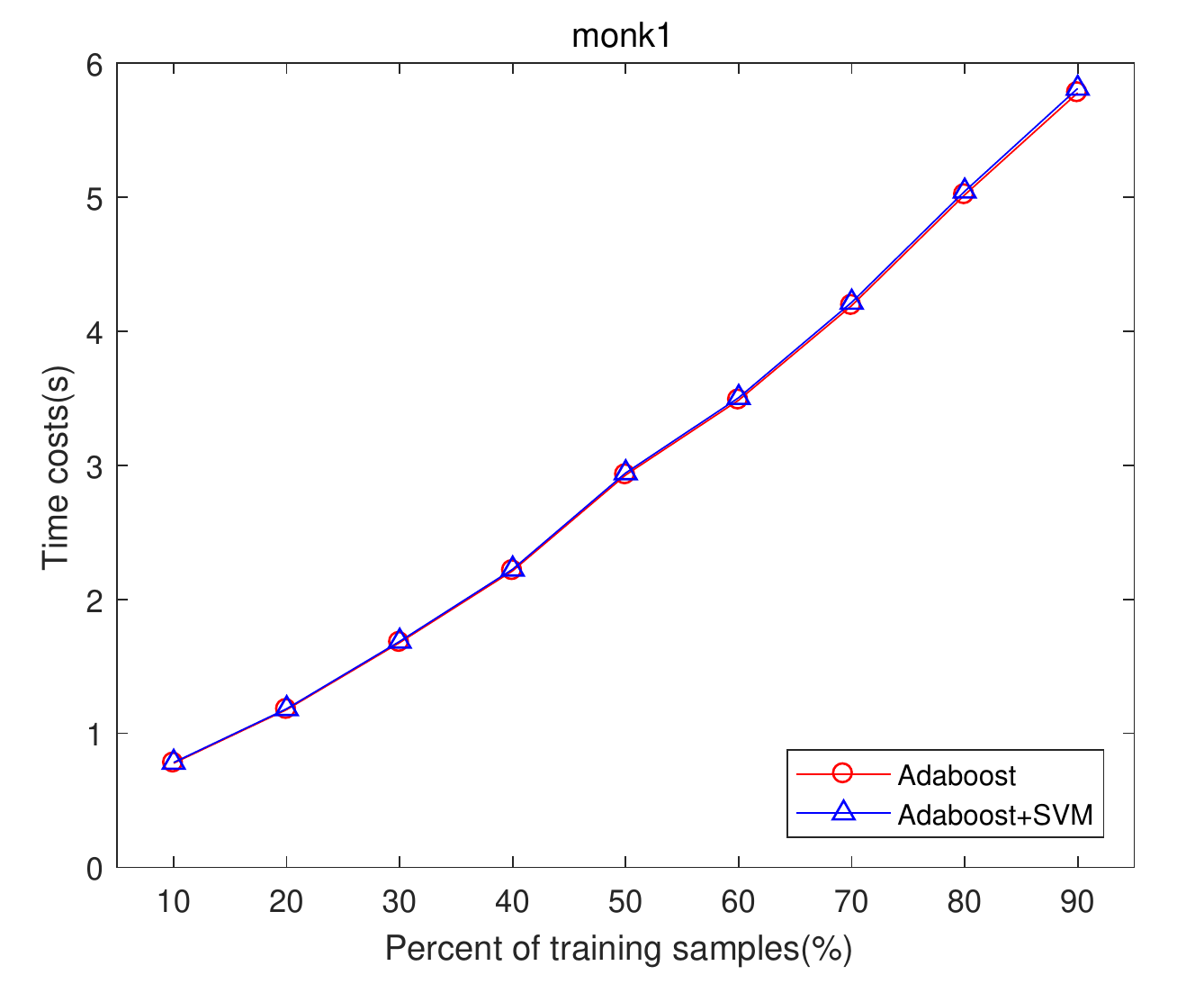}\label{Fig.8b}}\\
\caption{Classification accuracy ($\%$) and time costs vs the percent of labeled samples on monk1 dataset.\label{Fig.6}}
\end{figure}
\subsection{Datasets}
We utilize the following $4$ binary datasets to evaluate the performance of our model.
\begin{enumerate}
  \item fourclass: This dataset totally has $862$ samples and $2$ dimensions.
  \item ionosphere: This dataset is one of the UCI dataset with $351$ samples and $34$ features.
  \item chess: This dataset is also belongs to UCI dataset with $3196$ samples and $36$ features.
  \item monk1: This dataset is also one of the UCI dataset with $432$ samples and $6$ features.
\end{enumerate}
The detail descriptions of all datasets are also listed in Table~\ref{table.1}.
\begin{table}
\caption{Dataset Description}
\label{table.1}
\begin{center}
\resizebox{0.49\textwidth}{!}{
\renewcommand\arraystretch{1.25}
\begin{tabular}{p{25pt}p{25pt}<{\centering}p{25pt}<{\centering}||p{32pt}p{25pt}<{\centering}p{25pt}<{\centering}}
\hline
Dataset & Samples & Features & Dataset & Samples & Features  \\
\hline
fourclass & 862 & 2 &ionosphere & 351 & 34 \\
chess & 3196 & 36 & monk1 & 432 & 6 \\
\hline
\end{tabular}}
\end{center}
\end{table}
\subsection{Comparison Methods}
To demonstrate the effectiveness of the proposed approaches, we compare it with the classical AdaBoost algorithm. For all the methods, we run $5$ times and evaluate the classification results with the average classification accuracies and time costs. All the experiments are implemented in MATLAB R$2017$a, and run on a windows $10$ machine with a $3.6$GHz Intel i$7$-$7700$ CPU and $32$GB RAM.
\begin{table}[!ht]
  \centering
  \caption{Classification accuracy ($\%$) and time cost (s) vs the number of weak learner on fourclass dataset}
  \resizebox{0.49\textwidth}{!}{
  \renewcommand\arraystretch{1.25}
  \begin{tabular}{c|cc|cc}
    \hline\hline
    \multirow{2}{*}{Learner} & \multicolumn{2}{c|}{Classification accuracy ($\%$)} & \multicolumn{2}{c}{Time cost (s)} \cr
    &AdaBoost  & AdaBoost+SVM &AdaBoost  & AdaBoost+SVM \cr \hline \hline
    100  & 88.74 & 93.27 & 18.00 & 18.08  \\
    200  & 89.44 & 91.99 & 18.00 & 18.08 \\
    300  & 89.78 & 92.45& 30.45 & 30.57 \\
    400 & 89.55 & 92.58 & 47.82 & 48.00\\
    500  & 89.90 & 92.92& 65.63 & 65.88\\
    600 & 90.13 & 92.00 & 74.69 & 74.96 \\
    700& 90.37 & 92.69 & 113.43 & 113.85 \\
    800& 90.48 & 91.53 & 138.46 & 138.96  \\
    900& 90.60 & 91.30 & 163.72 & 164.27 \\
    1000& 89.91 & 90.60& 194.53 & 195.19  \\
    \hline \hline
  \end{tabular}}
  \label{table.2}
\end{table}

\begin{table}[!ht]
  \centering
  \caption{Classification accuracy ($\%$) and time cost (s) vs the number of weak learner on ionosphere dataset}
  \resizebox{0.49\textwidth}{!}{
  \renewcommand\arraystretch{1.25}
  \begin{tabular}{c|cc|cc}
    \hline\hline
    \multirow{2}{*}{Learner} & \multicolumn{2}{c|}{Classification accuracy ($\%$)} & \multicolumn{2}{c}{Time cost (s)} \cr
    &AdaBoost  & AdaBoost+SVM &AdaBoost  & AdaBoost+SVM \cr \hline \hline
    100  & 90.88 & 90.89 & 11.34 & 11.35  \\
    200  & 91.46 & 90.61 & 24.71 & 24.72 \\
    300  & 91.74 & 90.60& 36.10 & 36.12 \\
    400 & 91.74 & 90.89 & 47.51 & 47.54\\
    500  & 91.17 & 90.89& 60.66 & 60.69\\
    600 & 90.89 & 90.60 & 87.92 & 87.98 \\
    700& 91.45 & 90.32 & 104.49 & 104.56 \\
    800& 90.89 & 90.60 & 120.18 & 120.26  \\
    900& 91.45 & 90.32 & 135.99 & 136.08 \\
    1000& 90.89 & 90.60& 155.91 & 155.00  \\
    \hline \hline
  \end{tabular}}
  \label{table.3}
\end{table}

\begin{table}[!ht]
  \centering
  \caption{Classification accuracy ($\%$) and time cost (s) vs the number of weak learner on chess dataset}
  \resizebox{0.49\textwidth}{!}{
  \renewcommand\arraystretch{1.25}
  \begin{tabular}{c|cc|cc}
    \hline\hline
    \multirow{2}{*}{Learner} & \multicolumn{2}{c|}{Classification accuracy ($\%$)} & \multicolumn{2}{c}{Time cost (s)} \cr
    &AdaBoost  & AdaBoost+SVM &AdaBoost  & AdaBoost+SVM \cr \hline \hline
    100  & 95.87 & 97.09 & 545.83 &546.11  \\
    200  & 96.50 & 97.03 & 987.05 & 987.50 \\
    300  & 96.81 & 97.15& 1586.54 & 1587.39 \\
    400 & 96.81 & 97.18 & 2195.24 &2196.46\\
    500  & 96.93 & 97.28& 2551.54 & 2552.89\\
    600 & 96.93 & 97.31 & 3200.47 & 3202.18 \\
    700& 96.81 & 97.34 & 3873.00 & 3875.16 \\
    800& 96.75 & 97.37 & 4347.55 & 4349.90 \\
    900& 96.87 & 97.28 & 4807.12 & 4809.74 \\
    1000& 96.84 & 97.34& 5303.47 & 5306.41  \\
    \hline \hline
  \end{tabular}}
  \label{table.4}
\end{table}

\begin{table}[!ht]
  \centering
  \caption{Classification accuracy ($\%$) and time cost (s) vs the number of weak learner on monk1 dataset}
  \resizebox{0.49\textwidth}{!}{
  \renewcommand\arraystretch{1.25}
  \begin{tabular}{c|cc|cc}
    \hline\hline
    \multirow{2}{*}{Learner} & \multicolumn{2}{c|}{Classification accuracy ($\%$)} & \multicolumn{2}{c}{Time cost (s)} \cr
    &AdaBoost  & AdaBoost+SVM &AdaBoost  & AdaBoost+SVM \cr \hline \hline
    100  & 74.98 & 74.98 & 3.32 & 3.34  \\
    200  & 74.98 & 74.98 &6.95 & 6.99 \\
    300  & 74.98 & 74.98& 11.05 & 11.13 \\
    400 & 74.98 & 74.98 & 15.35 & 15.46\\
    500  & 74.98 & 74.98& 19.84 & 19.97\\
    600 & 74.98 & 74.98 & 24.75 & 24.92 \\
    700& 74.98 & 74.98 & 30.48 & 30.67 \\
    800& 74.98 & 74.98 & 88.88 & 88.98  \\
    900& 74.98 & 74.98 & 35.98 & 36.21 \\
    1000&74.98 & 74.98& 41.80 & 42.05  \\
    \hline \hline
  \end{tabular}}
  \label{table.5}
\end{table}

\subsection{The effect on the number of weak learner}
In most AdaBoost algorithm, the number of weak learner is set empirically. Now, we conduct experiments on the above four binary datasets to observe the effect on the number of weak learner. We adopt the $10$-fold cross validation way \cite{Kohavi:1995:SCB:1643031.1643047} to select $90\%$ in each class labeled samples as the training data to construct the classifier, the rest $10\%$ samples as the testing data to evaluate the performance of this classifier. Firstly, each dataset $S$ was divided into $10$ mutually exclusive subsets of similar size, that is $S=S_1 \bigcup S_2 \bigcup \cdots \bigcup S_{10}, S_i \bigcap S_j=\varnothing (i\neq j)$. Besides, each subset keeps the data distribution as consistent as possible. Then, the set of $9$ subsets was used as a training set, and the remaining subset as the test set in each time, which can obtain $10$ groups training and testing set. In this way, $10$ times training and testing procedures can be conducted. Finally, we can calculate the mean of $10$ times testing results.

The number of weak learner varies from $100$ to $1000$ and the corresponding classification accuracy ($\%$) and time cost (s) on the four datasets are recorded in Table~\ref{table.2}, Table~\ref{table.3}, Table~\ref{table.4} and Table~\ref{table.5}. From Table~\ref{table.2} to Table~\ref{table.5}, we have the following observation. Firstly, on the fourclass and chess datasets, the classification accuracies obtained by AdaBoost+SVM are a little higher than AdaBoost+SVM. On the monk1 dataset, the classification accuracies calculated by the two models are exactly equal. On the ionosphere dataset, the classification accuracies obtained by AdaBoost+SVM are a little lower than AdaBoost+SVM. However, the gap between the results of AdaBoost and AdaBoost+SVM is not very large. Then, we can see that time costs obtained by AdaBoost and AdaBoost+SVM have little gap. The reason is that SVM can quickly handle classification problems.  Therefore, we can conclude that our AdaBoost+SVM is very close to the AdaBoost, which can be regarded as a new explanation to the AdaBoost algorithm.
\subsection{The effect on the number of training data}
Next, we conduct experiments on the above datasets to observe the effect on the number of training data. For all the datasets, we fix the number of weak learner to $200$. We vary the percent of labeled samples in each class from $10\%$ to $90\%$ as the training data, the remaining samples as testing data. The results are shown in Fig.~\ref{Fig.3}, Fig.~\ref{Fig.4}, Fig.~\ref{Fig.5} and Fig.~\ref{Fig.6}.

As a general trend, the classification accuracy and time costs increase with the number of training samples increasing on all the datasets. However, more training data brings more time to train this classifier.
Then, in terms of classification accuracies and time costs on all the datasets, the results of AdaBoost and AdaBoost+SVM are still close, which verifies the performances of the two is comparable again.

\section{Conclusion}\label{sec.5}
In this paper, we have presented an AdaBoost+SVM model from the feature learning viewpoint to explain the success of AdaBoost that can resist the overfitting problem. Instead of directly weighted combination the base classifiers calculated by AdaBoost, we regard them as the new features and input them to the SVM classifier. The iterations increase means the dimensions of features are increase, so that the performance of SVM would be improved, which can explain the resistant overfitting phenomenon of AdaBoost model in a simple way. The results on four binary datasets show that AdaBoost+SVM can produce the comparable results to the AdaBoost algorithm, which illustrates the rationality to understand the AdaBoost algorithm from the feature learning viewpoint.

\bibliographystyle{IEEEtran}
\bibliography{IEEEabrv,mybibfile}

\end{document}